\documentclass[11pt]{article}

\usepackage[preprint]{acl}

\usepackage{times}
\usepackage{latexsym}
\usepackage[T1]{fontenc}
\usepackage[utf8]{inputenc}
\usepackage{microtype}
\usepackage{graphicx}
\usepackage{booktabs}
\usepackage{amsmath}
\usepackage{multirow}

\title{Beyond the Mean: Within-Model Reliable Change Detection for LLM Evaluation}

\author{Jon-Paul Cacioli \\
  Independent Researcher, Melbourne, Australia \\
  ORCID: 0009-0000-7054-2014 \\
  \texttt{github.com/synthiumjp/beyond\_the\_mean}}

\begin{document}
\maketitle

\begin{abstract}
We adapted the Reliable Change Index \cite[RCI;][]{jacobson1991clinical} from clinical psychology to item-level LLM version comparison on 2,000 MMLU-Pro items ($K = 10$ samples at $T = 0.7$). Two within-family pairs were tested: Llama~3 to 3.1 (+1.6 points) and Qwen~2.5 to 3 (+2.8 points). On the full benchmark, most items showed no reliable change (79\% and 72\%). However, over half the items were floor/ceiling. Among analysable items, change was bidirectional with large effect sizes: 34\% improved and 28\% deteriorated for Llama; 47\% improved and 39\% deteriorated for Qwen (median $|\Delta p| = 0.50$ and $0.90$). Churn was asymmetric by difficulty: low-accuracy items improved, high-accuracy items deteriorated. Domain-level decomposition revealed family-specific reversals: Llama lost physics while Qwen lost law. Greedy single-shot evaluation missed 42\% of reliably changed items and falsely flagged 25\% of unchanged items. The aggregate accuracy gain is the net residual of opposing item-level movements. We recommend reporting churn rate alongside aggregate accuracy.\footnote{Pre-registration: \url{https://osf.io/3dnsa}. Code and data: \url{https://github.com/synthiumjp/beyond_the_mean}.}
\end{abstract}

\section{Introduction}

Model-version comparison is the most routine evaluation in LLM development. A model family releases version $N+1$. Accuracy on a benchmark changes by some number of percentage points. The change is reported as improvement or regression. This is the standard practice.

The problem is not that aggregate comparisons lack statistical rigour. \citet{miller2024error} introduced confidence intervals and cluster-adjusted standard errors for aggregate eval scores. That contribution addresses sampling error at the test level. But aggregate statistical inference answers a test-level question: is the overall score difference distinguishable from sampling noise? It does not address what happened at the item level.

Evidence that item-level behaviour changes substantially across model versions already exists, though it has not been analysed with measurement-theoretic tools. \citet{chen2024chatgpt} documented dramatic performance drift in GPT-3.5 and GPT-4 over a three-month window, with GPT-4 accuracy on prime number identification dropping from 84\% to 51\% between March and June 2023. \citet{echterhoff2024muscle} formalised the concept of \emph{negative flips}: instances where a previously correct prediction becomes incorrect after a model update. They observed negative flips across multiple model pairs and proposed a compatibility adapter to reduce them by up to 40\%. These studies establish that version updates produce item-level regressions alongside aggregate gains. What is missing is a principled statistical framework for quantifying how much of the observed item-level change exceeds measurement noise.

A 2 percentage point aggregate improvement is compatible with at least three item-level patterns. First, a uniform small improvement across all items. Second, large improvement on a minority of items with no change on the majority. Third, improvement on some items, deterioration on others, with a net positive aggregate. These patterns have different implications for what the version update accomplished. The aggregate cannot distinguish them.

The Reliable Change Index \cite[RCI;][]{jacobson1991clinical} is the standard method in clinical psychology for determining whether a within-subject change exceeds measurement error. The index divides an observed score change by the standard error of the difference, derived from the instrument's test-retest reliability. If the ratio exceeds 1.96, the change is reliable at $p < .05$. The framework classifies individual patients as reliably improved, unchanged, or reliably deteriorated. \citet{maassen2004standard} provided corrections to the standard error formula. \citet{chalmers2025empirical} recently extended RCI with IRT-based empirical priors.

This study adapts reliable-change logic to stochastic LLM evaluation by treating repeated generations as measurement occasions and benchmark items as units of change. The adaptation is not a direct transplant of the clinical method. In therapy outcome research, the unit of analysis is a person with a stable latent trait measured by a multi-item instrument. Here, the unit is a benchmark item with stochastic accuracy estimated from $K$ repeated generations. The reliability coefficient captures stochastic stability of the model's response distribution rather than internal consistency of a psychological instrument. What is preserved is the core logic: a change score is meaningful only if it exceeds what measurement imprecision alone could produce.

\subsection{The gap this study addresses}

The emerging psychometric evaluation literature for LLMs includes several contributions at the aggregate or instrument level. \citet{truong2025fantastic} applied measurement-theoretic methods to benchmark quality assurance. Multiple groups have applied Item Response Theory to LLM evaluation \cite{lalor2024benchmarking,polo2024tinybenchmarks,zhuang2024static}. \citet{vendrow2025benchmarks} constructed platinum benchmarks by removing label errors. A systematic review of 445 benchmarks identified pervasive construct validity problems \cite{sievert2025measuring}. A comprehensive review of LLM psychometrics \cite{ye2026llm} covers reliability, validity, generalizability, and measurement invariance but does not mention clinical change detection methods.

On the engineering side, the negative-flip literature \cite{echterhoff2024muscle,yan2021positive,cai2022measuring} documents that model updates produce item-level regressions and proposes training-time mitigations. But negative-flip counting operates on single greedy evaluations. It treats each item as a binary correct/incorrect outcome without accounting for stochastic response variability. An item that flips from correct to incorrect on a single greedy trial may be within the model's noise band. An item that does not flip may nonetheless have shifted substantially in its underlying response probability. A reliability-grounded framework is needed to distinguish genuine change from stochastic artefact.

No study in the ML evaluation literature applies clinical change detection methods to model-version comparison. The RCI framework is absent from both the psychometric LLM literature and the negative-flip engineering literature.

\section{Method}

The study was pre-registered prior to data collection.\footnote{OSF: https://osf.io/3dnsa} All code and data are available.\footnote{https://github.com/synthiumjp/beyond\_the\_mean}

\subsection{Design}

Two within-family model-version comparisons on the same 2,000 MMLU-Pro items \cite{wang2024mmlu}: 500 per domain (physics, law, psychology, economics; 10 options A--J; seed~42). Pair~1 (minor update): Llama-3-8B-Instruct versus Llama-3.1-8B-Instruct. Pair~2 (generational update): Qwen-2.5-7B-Instruct versus Qwen3-8B in non-thinking mode.

Each item was administered to each model $K = 10$ times at $T = 0.7$ with independent conversation contexts and deterministic seeds (seed $=$ item\_index $\times$ 10 $+$ $k$). Total: 80,000 trials. All models were run at Q5\_K\_M quantisation on an AMD Radeon RX 7900 GRE (16\,GB VRAM) via llama-cpp-python with Vulkan backend.

\subsection{Measures}

\paragraph{Per-item accuracy.} For each item $i$ and model $m$: $p(i, m) =$ proportion correct across $K = 10$ samples. Range 0.0 to 1.0 in 0.1 increments.

\paragraph{Reliability.} Estimated via 1,000 random split-halves of $K = 10$ into two halves of 5. Spearman-Brown corrected. Median reported with 95\% CI. ICC (two-way random, single measures) reported as robustness check.

\paragraph{Standard error of the difference.} $\text{SEM} = \text{SD}(p) \times \sqrt{1 - r_{xx}}$, computed per model. $S_{\text{diff}} = \sqrt{\text{SEM}_{v1}^2 + \text{SEM}_{v2}^2}$, computed per pair.

\paragraph{Reliable Change Index.} $\text{RCI}(i) = (p(i, v_2) - p(i, v_1)) / S_{\text{diff}}$. Reliable improvement: $\text{RCI} > 1.96$. No reliable change: $|\text{RCI}| \leq 1.96$. Reliable deterioration: $\text{RCI} < -1.96$.

\paragraph{Item exclusion.} Items excluded if: (a)~fewer than 6 valid responses for either model, or (b)~per-item accuracy exactly 0.0 or 1.0 in both models (floor/ceiling items with undetectable change). All primary proportions reported before and after exclusion.

\section{Results}

Zero missing responses across all 80,000 trials. Zero thinking traces detected in Qwen3-8B responses.

\subsection{Reliability}

Reliability was high across all four models. Llama~3: $r_{xx} = .973$ [.969, .976]. Llama~3.1: $r_{xx} = .966$ [.962, .968]. Qwen~2.5: $r_{xx} = .988$ [.987, .990]. Qwen~3: $r_{xx} = .996$ [.995, .996]. ICC values were lower but consistent (range .735 to .958). Reliability at $K = 4$ already exceeded .93 for all models. Spearman-Brown extrapolation indicated that $K = 2$ suffices for $r_{xx} = .80$ and $K = 3$ for $r_{xx} = .90$. Table~\ref{tab:reliability} reports measurement parameters.

\begin{table}[t]
\centering
\small
\begin{tabular}{lcccccc}
\toprule
\textbf{Model} & $r_{xx}$ & ICC & SEM & $S_{\text{diff}}$ & Min $\Delta$ & Acc. \\
\midrule
Llama 3 & .973 & .780 & .058 & & & 37.0\% \\
Llama 3.1 & .966 & .735 & .062 & .085 & .167 & 38.6\% \\
Qwen 2.5 & .988 & .894 & .045 & & & 44.9\% \\
Qwen 3 & .996 & .958 & .031 & .055 & .107 & 47.7\% \\
\bottomrule
\end{tabular}
\caption{Reliability estimates and measurement parameters. $S_{\text{diff}}$ and minimum detectable delta reported per pair.}
\label{tab:reliability}
\end{table}

\subsection{Exclusions}

The Llama pair excluded 1,048 items (52.4\%) due to floor/ceiling: 906 items were always wrong and 524 always correct in Llama~3. The Qwen pair excluded 1,348 items (67.4\%): 953 always wrong and 763 always correct in Qwen~2.5. Zero items were excluded for insufficient valid responses. The high exclusion rate reflects that at the 7--8 billion parameter scale, most MMLU-Pro items are either trivially correct or consistently intractable. The analysable items are those in the middle difficulty range where stochastic variation occurs.

\subsection{RCI classification}

Across the full 2,000-item benchmark, the pre-registered primary hypothesis (H1) was supported. For Llama, 78.7\% of items showed no reliable change. For Qwen, 72.0\%. On the full benchmark, reliable improvement (11.0\% Llama, 15.3\% Qwen) and reliable deterioration (10.3\% Llama, 12.7\% Qwen) each affected roughly one in ten items. Full-benchmark churn rates were 21.3\% (Llama) and 28.0\% (Qwen).

Within the analysable subset, the pattern shifted. For Llama ($n = 952$), 33.7\% reliably improved, 37.9\% showed no reliable change, and 28.4\% reliably deteriorated. For Qwen ($n = 652$), 46.9\% reliably improved, 14.1\% showed no change, and 39.0\% reliably deteriorated. The net surplus producing the aggregate gain is small: 51 items for Llama, 52 for Qwen. Table~\ref{tab:rci} reports the post-exclusion classification.

\begin{table}[t]
\centering
\small
\begin{tabular}{lccccc}
\toprule
\textbf{Pair} & $n$ & Impr. & No ch. & Deter. & Churn \\
\midrule
Llama & 952 & 33.7\% & 37.9\% & 28.4\% & 44.7\% \\
Qwen & 652 & 46.9\% & 14.1\% & 39.0\% & 85.9\% \\
\bottomrule
\end{tabular}
\caption{RCI classification of analysable items (post-exclusion). Churn = proportion reliably changed in either direction.}
\label{tab:rci}
\end{table}

\subsection{Effect sizes}

Across all analysable items (not conditional on RCI classification), the mean $|\Delta p|$ was 0.31 for Llama ($n = 952$) and 0.67 for Qwen ($n = 652$). 56.4\% of Llama analysable items and 83.7\% of Qwen analysable items showed $|\Delta p| \geq 0.2$. The distributions are wide. The changes are not concentrated at the detection threshold.

\begin{figure}[t]
\centering
\includegraphics[width=\columnwidth]{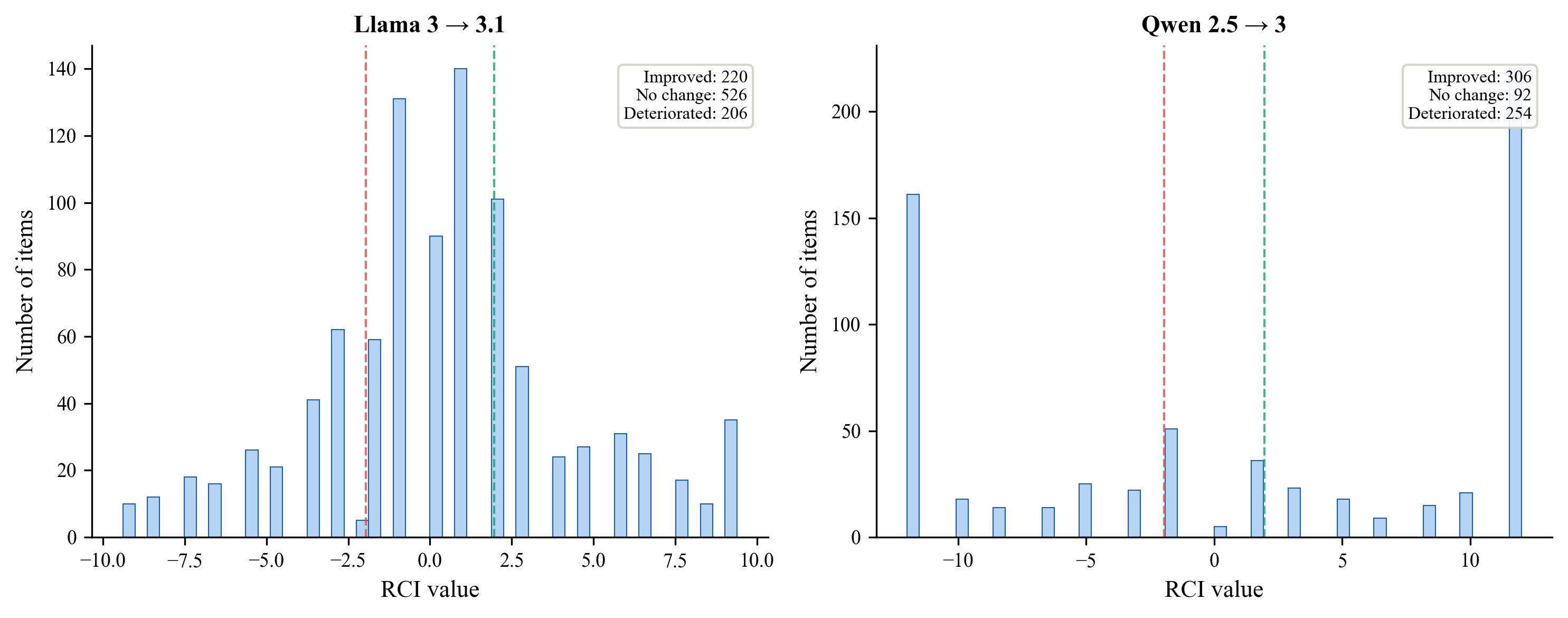}
\caption{RCI value distribution for both model pairs (post-exclusion). Dashed lines mark $|\text{RCI}| = 1.96$. Llama shows values clustered near the threshold; Qwen shows values spread across a wider range.}
\label{fig:rci_dist}
\end{figure}

Among the subset classified as reliably changed, effect sizes were larger. For Llama ($n = 426$), the mean $|\Delta p|$ was 0.56 and the median was 0.50. 70.7\% shifted by 0.4 or more. For Qwen ($n = 560$), the mean $|\Delta p|$ was 0.77 and the median was 0.90. 84.3\% shifted by 0.4 or more. These are large probability shifts representing 5--9 additional correct answers out of 10 samples. Figure~\ref{fig:rci_dist} displays the RCI magnitude distribution for both pairs.

\subsection{Churn by baseline difficulty}

\begin{figure}[t]
\centering
\includegraphics[width=\columnwidth]{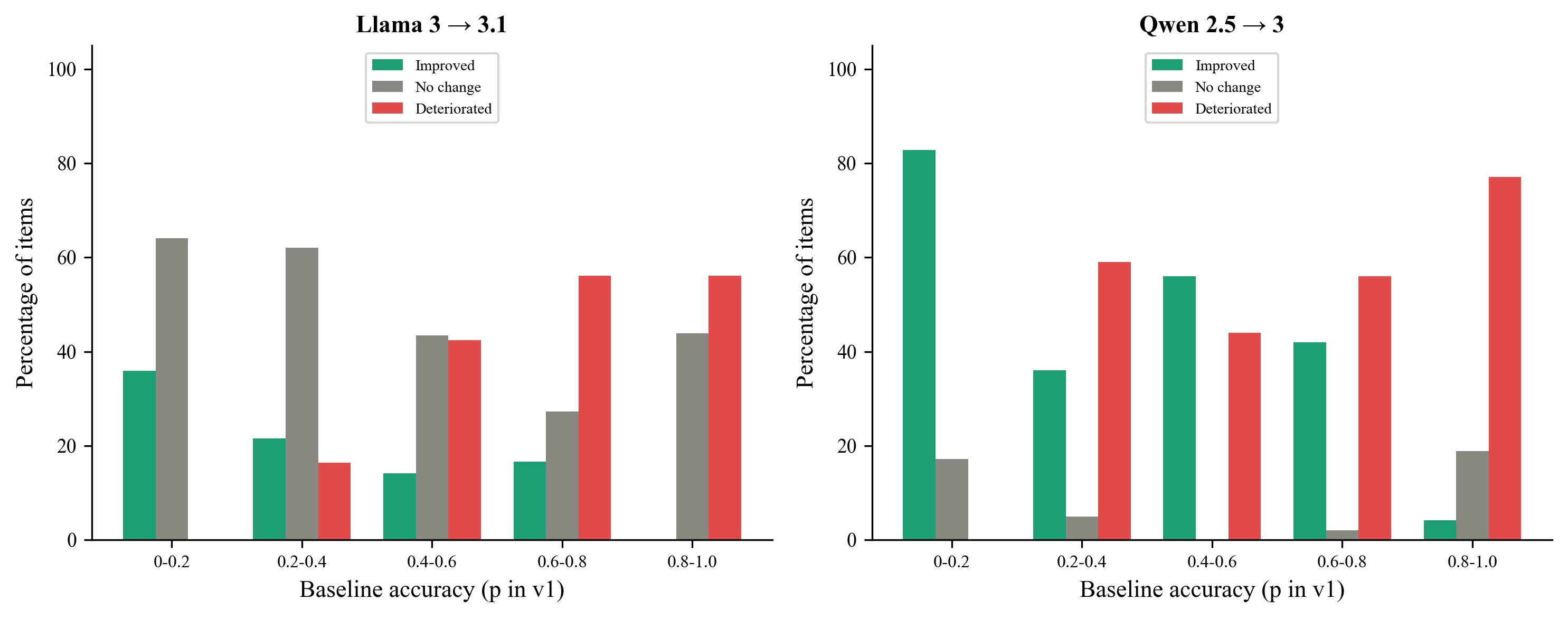}
\caption{Churn by baseline difficulty ($p$ in $v_1$). Low-accuracy items predominantly improve; high-accuracy items predominantly deteriorate.}
\label{fig:difficulty}
\end{figure}

Churn was asymmetric by baseline difficulty. For Llama, items with low baseline accuracy ($p_{v1} = 0.0$--$0.2$, $n = 454$) showed 35.9\% churn, almost entirely improvement (163 improved, 0 deteriorated). Items with high baseline accuracy ($p_{v1} = 0.8$--$1.0$, $n = 187$) showed 56.1\% churn, almost entirely deterioration (0 improved, 105 deteriorated). Middle-difficulty items showed mixed bidirectional change. Figure~\ref{fig:difficulty} displays the full pattern.

The pattern for Qwen was more extreme. Low-accuracy items ($n = 273$): 82.8\% churn, all improvement. High-accuracy items ($n = 218$): 81.2\% churn, predominantly deterioration. Middle bins showed near-complete churn (95--100\%) with bidirectional changes. In both pairs, items near the floor of performance tended to improve while items near the ceiling tended to deteriorate, consistent with regression toward the middle of the accuracy range overlaid on domain-specific version effects.

\subsection{Empirical null calibration}

The empirical null (1,000 block-level permutations) confirmed that observed reliable-change counts exceeded chance. For Llama, observed improvement (321) exceeded the null 95th percentile (229). Observed deterioration (270) also exceeded the null (229). For Qwen, observed improvement (306) exceeded the null (299). Observed deterioration (254) did not exceed the null (299). The Qwen deterioration count is within what version-label permutation could produce.

\subsection{Domain analysis}

\begin{figure}[t]
\centering
\includegraphics[width=\columnwidth]{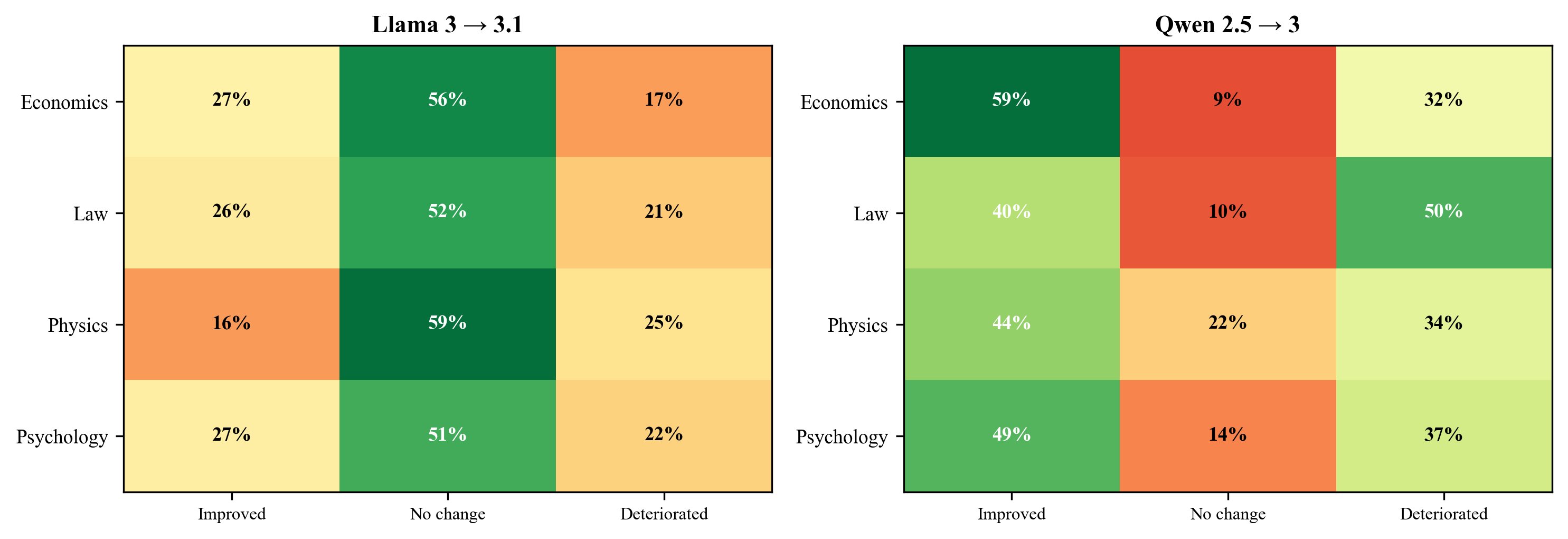}
\caption{Domain $\times$ RCI category heatmap (post-exclusion). Llama loses physics; Qwen loses law. The domains that deteriorate differ across families.}
\label{fig:domain}
\end{figure}

Domain clustering (H2) was supported for both pairs (Llama: $\chi^2 = 26.5$, $p < .001$, Cram\'{e}r's $V = .118$; Qwen: $\chi^2 = 28.6$, $p < .001$, $V = .148$). Llama~3.1 gained in economics, law, and psychology but deteriorated in physics (improvement/deterioration ratio $= 0.66$). Qwen~3 gained in economics, physics, and psychology but deteriorated in law (ratio $= 0.80$). The domains that gain and lose differ across model families. Figure~\ref{fig:domain} displays the domain-level heatmap.

\subsection{Cross-pair analyses}

H3 (cross-pair divergence) was supported: Qwen showed a higher reliable-change rate (85.9\%) than Llama (62.1\%), $z = -10.4$, $p < .001$. This is consistent with the generational versus minor update distinction.

The cross-pair item-level RCI correlation was near zero ($r = .11$, $p = .019$, $n = 431$ shared analysable items). The items that changed reliably for Llama are almost entirely different from those that changed for Qwen.

\subsection{Greedy comparison}

Comparison with greedy ($T = 0$) data for the Llama pair showed 86.6\% exact agreement between greedy binary changes and RCI classifications across 1,997 matched items. Of 329 items that greedy scoring flagged as changed, 83 (25.2\%) were classified as no reliable change by RCI. Of 425 items that RCI classified as reliably changed, 179 (42.1\%) showed no change under greedy scoring. Single-shot greedy evaluation both overdetects noise-driven flips and underdetects genuine probability shifts.

\subsection{Item-level scatter}

\begin{figure*}[t]
\centering
\includegraphics[width=0.48\textwidth]{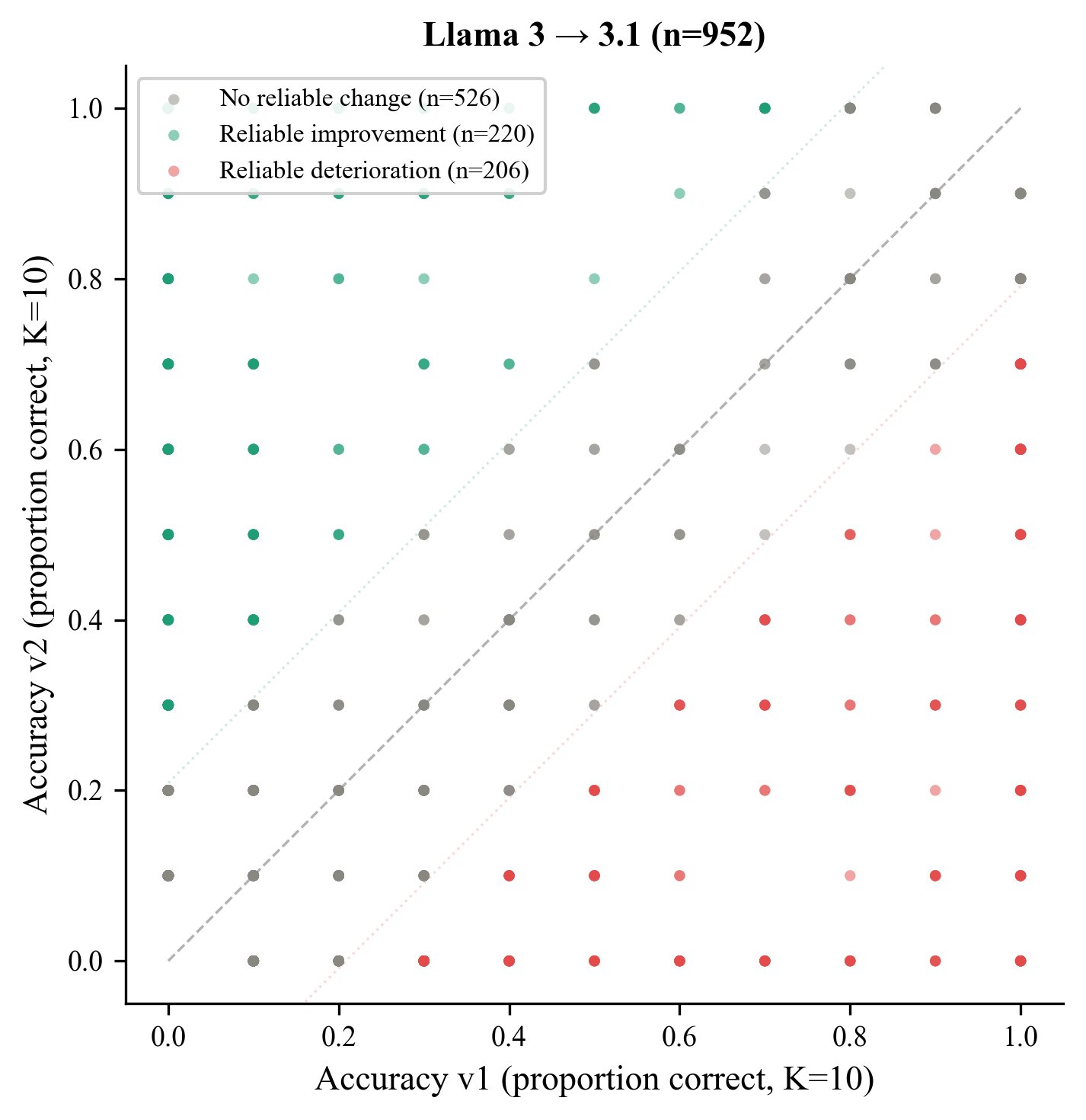}
\hfill
\includegraphics[width=0.48\textwidth]{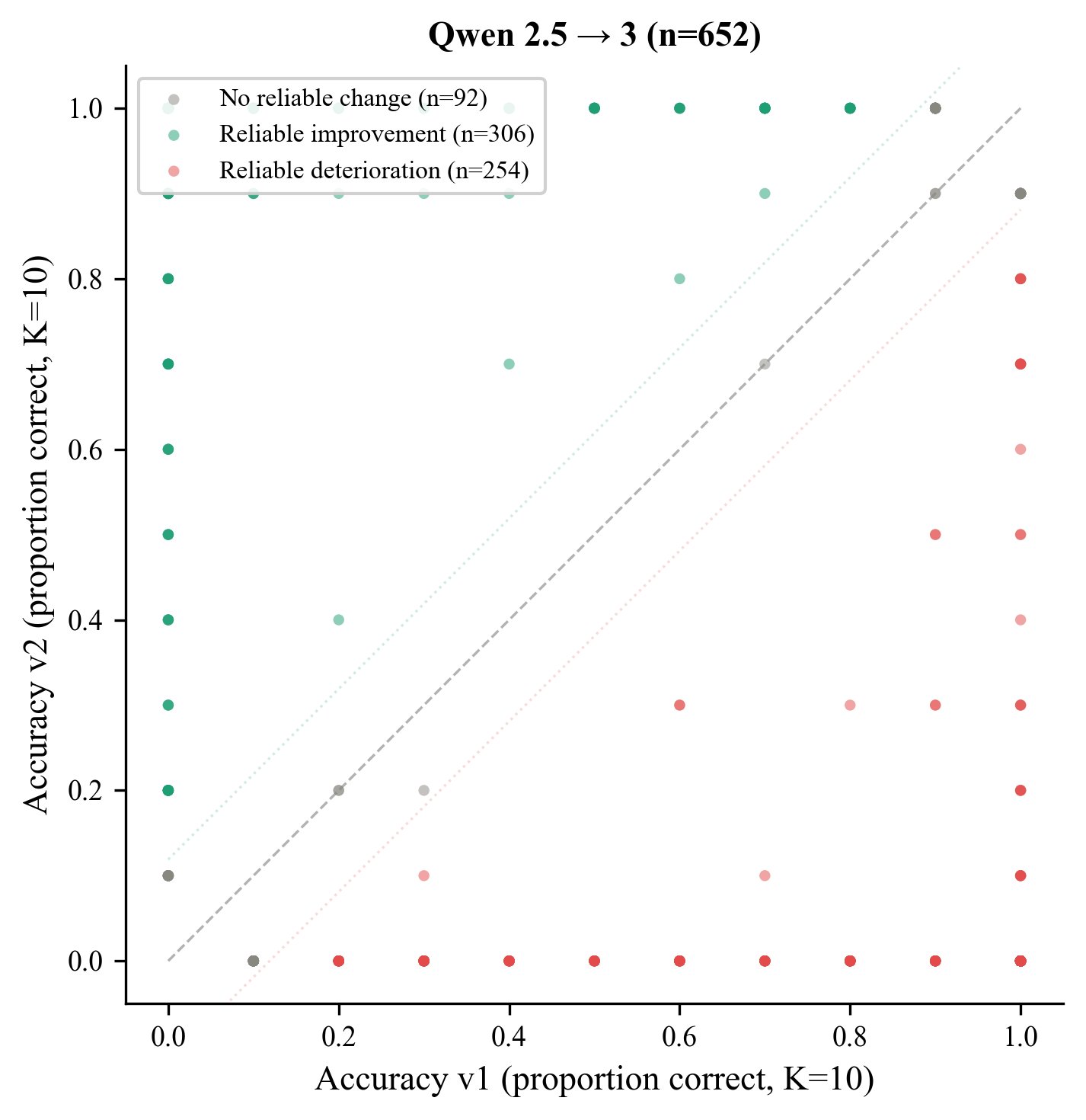}
\caption{Per-item accuracy in $v_1$ versus $v_2$, coloured by RCI classification. Left: Llama 3 $\to$ 3.1 ($n = 952$). Right: Qwen 2.5 $\to$ 3 ($n = 652$). Dashed line: no change. Dotted lines: RCI detection band.}
\label{fig:scatter}
\end{figure*}

Figure~\ref{fig:scatter} displays per-item accuracy in $v_1$ versus $v_2$ for both model pairs. Items above the diagonal improved; items below deteriorated. The colour coding shows that reliable changes (green and red) cluster away from the diagonal while no-change items (grey) cluster near it.

\section{Discussion}

\subsection{Two levels of the finding}

The results operate at two levels. On the full benchmark, the pre-registered prediction held. Most items showed no reliable change (78.7\% for Llama, 72.0\% for Qwen). This result is dominated by floor and ceiling items that are deterministically correct or incorrect regardless of model version. On the analysable subset where stochastic variation occurred, bidirectional churn with large effect sizes was the dominant pattern.

The distinction matters. The churn finding is conditional on the analysable subset. It does not mean the entire benchmark is in flux. It means that items where change can be detected show a pattern that aggregate reporting conceals. The net gain is the small residual of two large opposing movements.

The effect sizes among reliably changed items were not trivially small. For Llama, the median $|\Delta p|$ was 0.50. For Qwen, the median was 0.90. These represent shifts of 5--9 correct answers out of 10 samples.

\subsection{Regression toward the middle}

The difficulty-bin analysis revealed a clear asymmetric pattern. Items with low baseline accuracy predominantly improved. Items with high baseline accuracy predominantly deteriorated. Middle-difficulty items showed mixed bidirectional change. This is consistent with regression toward the middle of the accuracy range. Items near the performance floor can only go up. Items near the ceiling can only go down.

Regression toward the middle explains the directional asymmetry within difficulty bins. It does not explain the domain-level patterns. Llama~3.1 lost ground in physics while gaining in economics, law, and psychology. Qwen~3 lost ground in law while gaining in economics, physics, and psychology. The domains that deteriorate differ across model families. Regression dynamics are domain-blind. If the bidirectional churn were driven entirely by regression, the same domains would show the same directional patterns in both pairs. They do not. The domain-specific reversals indicate structured, family-specific effects overlaid on the regression baseline.

Three mechanisms therefore contribute to the observed churn. First, regression toward the middle, which is predictable from baseline difficulty and accounts for the directional asymmetry at the extremes. Second, domain-specific training effects, which produce the family-specific reversal pattern. Third, stochastic measurement noise, which the empirical null calibration addresses. The empirical null confirmed that observed reliable-improvement counts exceeded chance for both pairs. The domain-level patterns survived chi-squared testing with $p < .001$ for both pairs. The regression component is real but it is not the whole story.

\subsection{Domain-specific patterns}

Llama~3.1 gained in three domains and lost in physics. Qwen~3 gained in three domains and lost in law. The domains that gain and lose differ across model families. In these data, a user who selected Llama~3.1 based on aggregate improvement would experience worse physics performance than Llama~3. A user who selected Qwen~3 would experience worse law performance than Qwen~2.5. Neither user would know this from the aggregate.

\subsection{Single-shot evaluation}

The greedy comparison showed that standard single-shot binary evaluation misses 42\% of reliably changed items and falsely flags 25\% of unchanged items. The items it misses are those where the underlying response probability shifted but did not cross the binary correct/incorrect threshold. The items it falsely flags are those where a binary flip occurred within stochastic noise. $K = 3$ samples suffice for $r_{xx} = .90$. The computational overhead for reliability-informed comparison is minimal.

\subsection{Relationship to prior work}

\citet{miller2024error} brought inferential statistics to aggregate eval scores. His contribution addresses test-level precision. Ours addresses item-level decomposition. Both are needed. A therapy trial can show a statistically significant group-level improvement while individual patient analysis reveals that only a minority improved. The group-level finding is real but incomplete.

The negative-flip literature \cite{echterhoff2024muscle,yan2021positive} is the closest prior work. MUSCLE documented that model updates produce negative flips and proposed a compatibility adapter that reduced them by up to 40\%. Our contribution differs in three ways. First, negative-flip counting uses single greedy evaluations and treats each item as binary. Our greedy comparison showed that this misses 42\% of reliably changed items and falsely flags 25\% of unchanged items. Second, MUSCLE frames the problem as an engineering defect to mitigate during training. We frame it as a measurement phenomenon to quantify during evaluation. Third, MUSCLE does not report bidirectional structure. Our decomposition reveals that improvements and deteriorations occur in comparable proportions.

\citet{chen2024chatgpt} documented performance drift in GPT-3.5 and GPT-4, with GPT-4 accuracy on prime number identification dropping from 84\% to 51\% over three months. Their finding is reported at the aggregate task level. RCI would decompose such changes into item-level categories, distinguishing genuine shifts from noise.

\citet{truong2025fantastic} applied measurement theory to benchmark quality assurance. Their work identifies bad items. Ours characterises how items behave differently across model versions.

\section{Limitations}

\paragraph{High exclusion rate.} Over half the items were excluded as floor/ceiling. The analysable items are those in the middle difficulty range. Whether the churn finding extends to the full difficulty range cannot be determined from this design. A harder benchmark with fewer floor/ceiling items would have a larger analysable set.

\paragraph{Quantisation.} All models were run at Q5\_K\_M. Whether per-item accuracy under sampling is stable across quantisation formats is an open question.

\paragraph{Two model pairs.} Two pairs provide replication but not generalisation. Whether the churn finding extends to frontier-scale models, different benchmarks, or different version-update magnitudes requires further work.

\paragraph{Temperature dependence.} Results are conditional on $T = 0.7$. Different temperatures would produce different reliability estimates and potentially different RCI classifications.

\paragraph{Homoscedastic SEM.} The primary RCI uses a single $S_{\text{diff}}$ per pair. The pre-registered stratified sensitivity analysis showed that difficulty-stratified $S_{\text{diff}}$ produces even more reliable change, suggesting the global analysis is conservative.

\paragraph{Adaptation of RCI.} The RCI was designed for continuous scores from multi-item instruments measuring stable latent traits. Our adaptation applies it to binomial proportions from repeated stochastic samples. This adaptation preserves the core logic (change must exceed measurement noise) but the distributional assumptions differ. A beta-binomial or hierarchical Bayesian model would address the heteroscedasticity inherent in binomial data. We position RCI as a computationally simple approximation that is sufficient to reveal the bidirectional churn pattern.

\section{Conclusion}

In these data, at the 7--8 billion parameter scale, aggregate accuracy gains from model-version updates were the net residual of opposing item-level movements. On the full benchmark, most items showed no reliable change (79\% for Llama, 72\% for Qwen). Among analysable items where stochastic variation occurred, change was predominantly bidirectional, with large effect sizes (median $|\Delta p| = 0.50$--$0.90$). Full-benchmark churn rates were 21\% (minor update) and 28\% (generational update). The domains that gained and lost differed across model families. Single-shot evaluation misses 42\% of reliably changed items. The Reliable Change Index provides a principled, computationally inexpensive decomposition that reveals structure within aggregate accuracy differences. Whether these patterns generalise to frontier-scale models, other benchmarks, or different evaluation protocols is an open question. We recommend reporting churn rate alongside aggregate accuracy for version comparisons, and using $K \geq 3$ stochastic samples as a low-cost alternative to single-shot binary evaluation.

\bibliography{references}

@article{jacobson1991clinical,
  title={Clinical significance: A statistical approach to defining meaningful change in psychotherapy research},
  author={Jacobson, Neil S and Truax, Paula},
  journal={Journal of Consulting and Clinical Psychology},
  volume={59},
  number={1},
  pages={12--19},
  year={1991},
  publisher={American Psychological Association}
}

@article{maassen2004standard,
  title={The standard error in the {J}acobson and {T}ruax {R}eliable {C}hange {I}ndex: The classical approach to the assessment of reliable change},
  author={Maassen, Gerard H},
  journal={Journal of the International Neuropsychological Society},
  volume={10},
  number={6},
  pages={888--893},
  year={2004},
  publisher={Cambridge University Press}
}

@article{chalmers2025empirical,
  title={Including empirical prior information in the {R}eliable {C}hange {I}ndex},
  author={Chalmers, R Philip and Campbell, Sarah},
  journal={Applied Psychological Measurement},
  year={2025},
  doi={10.1177/01466216251358492},
  publisher={Sage}
}

@article{miller2024error,
  title={Adding error bars to evals: A statistical approach to language model evaluations},
  author={Miller, Evan},
  journal={arXiv preprint arXiv:2411.00640},
  year={2024}
}

@inproceedings{echterhoff2024muscle,
  title={{MUSCLE}: A Model Update Strategy for Compatible {LLM} Evolution},
  author={Echterhoff, Jessica and Faghri, Fartash and Vemulapalli, Raviteja and Hu, Ting-Yao and Li, Chun-Liang and Tuzel, Oncel and Pouransari, Hadi},
  booktitle={Findings of the Association for Computational Linguistics: EMNLP 2024},
  year={2024}
}

@article{chen2024chatgpt,
  title={How is {ChatGPT}'s behavior changing over time?},
  author={Chen, Lingjiao and Zaharia, Matei and Zou, James},
  journal={Harvard Data Science Review},
  volume={6},
  number={2},
  year={2024}
}

@inproceedings{yan2021positive,
  title={Positive-congruent training: Towards regression-free model updates},
  author={Yan, Sijie and Xiong, Yuanjun and Grover, Aditya and Sohn, Kihyuk and Chandraker, Manmohan and Jaakkola, Tommi},
  booktitle={Proceedings of the IEEE/CVF Conference on Computer Vision and Pattern Recognition},
  pages={14299--14308},
  year={2021}
}

@article{cai2022measuring,
  title={Measuring backward compatibility in code generation models},
  author={Cai, Tongshuang and Vasconcelos, Nuno and Dvijotham, Krishnamurthy},
  journal={arXiv preprint arXiv:2211.07889},
  year={2022}
}

@inproceedings{truong2025fantastic,
  title={Fantastic bugs and how to squash them},
  author={Truong, Son and Domingue, Benjamin and Koyejo, Sanmi},
  booktitle={Advances in Neural Information Processing Systems},
  year={2025}
}

@article{ye2026llm,
  title={Large language model psychometrics: A systematic review of evaluation, validation, and enhancement},
  author={Ye, Jingwei and Zhou, Junyi and Liu, Zilu and Hu, Jiaheng and Zhu, Mengyu and Li, Kunlun and Zhao, Hanjia and Xu, Jie},
  journal={arXiv preprint arXiv:2505.08245},
  year={2026}
}

@article{vendrow2025benchmarks,
  title={Do large language model benchmarks test reliability?},
  author={Vendrow, Joshua and Gu, Edward and Papadimitriou, Isabel and Hsu, Daniel and Hashimoto, Tatsunori},
  journal={arXiv preprint arXiv:2502.03461},
  year={2025}
}

@inproceedings{sievert2025measuring,
  title={Measuring what matters: Construct validity in large language model benchmarks},
  author={Sievert, Scott and Gao, Leo and Dorr, Bonnie and Hovy, Eduard},
  booktitle={Advances in Neural Information Processing Systems},
  year={2025}
}

@article{lalor2024benchmarking,
  title={Benchmarking large language models with item response theory},
  author={Lalor, John P and Yang, Yi},
  journal={Transactions of the Association for Computational Linguistics},
  year={2024}
}

@inproceedings{polo2024tinybenchmarks,
  title={tiny{B}enchmarks: Evaluating {LLM}s with fewer examples},
  author={Polo, Felipe Maia and Weber, Lucas and Choshen, Leshem and Sun, Yuekai and Xu, Gongjun and Yurochkin, Mikhail},
  booktitle={Proceedings of the International Conference on Machine Learning},
  year={2024}
}

@article{zhuang2024static,
  title={From static benchmarks to adaptive testing: Psychometrics in {AI} evaluation},
  author={Zhuang, Yuqing and Qi, Haoming and Li, Yifan and Jin, Di},
  journal={arXiv preprint arXiv:2306.10512},
  year={2024}
}

@inproceedings{wang2024mmlu,
  title={{MMLU-Pro}: A more robust and challenging multi-task language understanding benchmark},
  author={Wang, Yubo and Ma, Xueguang and Zhang, Ge and Ni, Yuansheng and Chandra, Abhranil and Guo, Shiguang and Ren, Weiming and Arulraj, Aaran and He, Xuan and Jiang, Ziyan and Li, Tianle and Ku, Max and Wang, Kai and Zhuang, Alex and Fan, Rongqi and Yue, Xiang and Chen, Wenhu},
  booktitle={Advances in Neural Information Processing Systems},
  year={2024}
}

\end{document}